\documentclass[runningheads]{llncs}

\usepackage{graphicx}
\usepackage{comment}
\usepackage{amsmath,amssymb}
\usepackage{color}
\usepackage{url}
\usepackage{hyperref}

\usepackage{csquotes}
\usepackage{enumitem}
\usepackage{subcaption}
\usepackage{cleveref}
\usepackage{array}

\newif\ifreview
 \reviewfalse

\ifreview
	\usepackage{lineno}

	\linenumbers
\fi

\makeatletter
\newcommand{\printfnsymbol}[1]{%
	\textsuperscript{\@fnsymbol{#1}}%
}
\makeatother

\begin{document}


\def\SubNumber{000}

\def\GCPRTrack{Regular Track}

\title{Paint it Black: Generating \\ paintings from text descriptions}

\ifreview
	\titlerunning{DAGM GCPR 2021 Submission \SubNumber{}. CONFIDENTIAL REVIEW COPY.}
	\authorrunning{DAGM GCPR 2021 Submission \SubNumber{}. CONFIDENTIAL REVIEW COPY.}
	\author{DAGM GCPR 2021 - \GCPRTrack{}}
	\institute{Paper ID \SubNumber}
\else

	\author{Mahnoor Shahid\inst{1} \and Mark Koch\inst{2} \and Niklas Schneider\inst{3}
     \\ \\
    \email{mash00001@uni-saarland.de}\inst{1} 
    \email{s8mkkoch@uni-saarland.de}\inst{2} \\
    \email{s8nlschn@uni-saarland.de}\inst{3} }

	\authorrunning{M. Shahid et al.}

	
	\institute{Universität des Saarlandes}

\maketitle              

\begin{abstract}
Two distinct tasks - generating photorealistic pictures from given text prompts and transferring the style of a painting to a real image to make it appear as though it were done by an artist, have been addressed many times, and several approaches have been proposed to accomplish them.
However, the intersection of these two, i.e., generating paintings from a given caption, is a relatively unexplored area with little data available. In this paper, we have explored two distinct strategies and have integrated them together. First strategy is to generate photorealistic images and then apply style transfer and the second strategy is to train an image generation model on real images with captions and then fine-tune it on captioned paintings later. 
These two models are evaluated using different metrics as well as a user study is conducted to get human feedback on the produced results.

\keywords{Text2Image  \and Style Transfer \and StyleGAN \and SSA-GAN}
\end{abstract}

\begin{figure}[h]
	\begin{center}
		\scriptsize
		\setlength{\tabcolsep}{1em}
		\makebox[\textwidth][c]{\begin{tabular}{p{3cm}p{3cm}p{3cm}p{3cm}}
			\includegraphics[width=3cm]{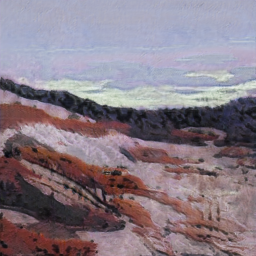}
			& \includegraphics[width=3cm]{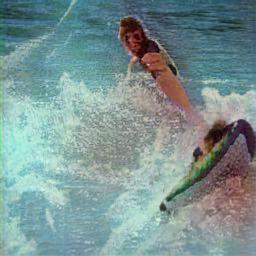}
			&\includegraphics[width=3cm]{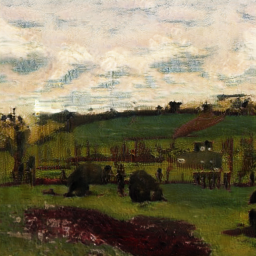}
			&\includegraphics[width=3cm]{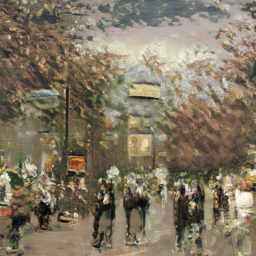}\\[3pt]
			\centering\ttfamily A snowy landscape in front of a mountain range
			&\centering\ttfamily A woman looks to the right while standing on a surfboard in the ocean
			&\centering\ttfamily Some cows standing on a lush green field on a cloudy day eating grass
			&\centering\ttfamily A group of people walking down a busy road
		\end{tabular}}
	\end{center}
	\vspace{-0.25cm}
	\caption{Paintings generated from text prompts using our work}
	\label{fig:demo}
\end{figure}
\vspace{-1cm}

\section{Introduction}

Synthetic art generation has been catching increasing interest in the art community over the last years with artists like Mario Klingemann, Trevor Paglen or Gene Kogan presenting exhibitions in famous galleries featuring paintings generated by artificial intelligence.\footnote{See for example Klingemann's exhibitions \enquote{X Degrees of Separation} at the Metropolitan Museum of Art in New York, or \enquote{Neural Decay} at the Centre Pompidou in Paris.}
In the future, such systems could become a standard technique in an artist's toolbox, possibly helping to find inspiration or aiding in the painting process.

In this work, we want to investigate different approaches to control the content of synthesized artworks by conditioning on text inputs.
That is, given a single-sentence natural language text prompt, we want to generate a painting that depicts the described scene (for example like in \Cref{fig:demo}).
Linguistic descriptions are well suited for our purpose since text is one of the most convenient mediums for humans to communicate and express visual ideas.

A general challenge in the area of text to image synthesis (T2I) is the fact that the problem is highly multimodal, meaning that there are many different configurations of pixels that can accurately capture a text description.
For example, you can probably find more than one image in a painting database matching to the description \enquote{a flowery landscape around a river}.
Luckily, deep learning methods have made incredible progress in those areas over the last years, with architectures offering reasonable results for various datasets.
The most widely adopted datasets in T2I research are CUB-200 Birds \cite{reed2016generative,wah2011cub}, Oxford 102 Flowers \cite{Nilsback2008flowers}, and COCO \cite{coco}.
The main difficulty in applying those existing T2I approaches to paintings is the availability of data:
Deep learning methods for T2I require large amounts of images that are already annotated with captions.
Since this kind of data is not freely available for paintings, solving this task becomes more challenging.
In this work, we investigated two different approaches to overcome this issue:
\begin{enumerate}[itemsep=10pt]
	\item
	We use an existing T2I architecture to generate photo-realistic images and turn them into paintings in a separate stylization step.
	
	\item 
	We generate our own captioned paintings dataset based on the WikiArt dataset and use this data to finetune a T2I model.
\end{enumerate}
We evaluate the generated images based on quality and caption accuracy using standard metrics and a human survey.

\section{Related Work}

Art synthesis has been a long-standing area of research interest, particularly in the deep learning community.
One of the most well-known approaches is called neural style transfer \cite{gatys2015style}, where a convolutional neural network is used to transport the artistic style of a painting onto another image.
This method can create artworks of high quality and also scales to very high resolutions.
However, the generation is completely based on the input images and cannot produce any novel content.

Generative adverserial networks (GANs) \cite{goodfellow2014GAN} on the other hand can synthesize completely new images, with modern architectures like StyleGAN being able to generate photorealistic images \cite{stylegan}.
Reed et al. \cite{reed2016generative} were the first to use GANs for T2I tasks by conditioning the generation process on a sentence embedding obtained from a pre-trained text-encoder.
They also ask the discriminator to detect mismatched captions to force the generator to produce images fitting to the text prompt.

Stacked architectures were introduced to allow T2I models to output higher resolutions.
For example, StackGAN \cite{zhang2017StackGAN} and StackGAN++ \cite{zhang2017StackGANPP} use multiple generators and discriminators of increasing resolutions.
AttnGAN \cite{xu2018AttnGAN} builds on StackGAN++ by incorporating an attention mechanism to focus on relevant words in the caption and conditioning different image regions on different words.
They also proposed the Deep Attentional Multimodal Similarity Model (DAMSM) loss that measures the similarity between the generated images and the input text.

DF-GAN \cite{tao2020DFGAN} introduced a simplified T2I backbone that is able to synthesize high-resolution images without the stacking approach, thus cutting down on training time.
They also employ affine transformations on the image feature maps whose scaling and shifting parameters are conditioned on the text features.
This way, image and text information is fused during the generation process.

While most T2I approaches are based on GANs, recently, there has also been work on using large transformers for this task.
For example, DALL-E \cite{ramesh2021dalle} is a version of GPT-3 \cite{brown2020gpt3} trained for T2I and produces very promising results.
However, because of the large computational costs this approach is not feasible for us.

\section{Methods}
In the this section we discuss the baseline model and the T2I architecture we employed and finally elaborate on our two proposed approaches.

\subsection{Baseline}

\begin{figure}[t]
	\begin{center}
		\includegraphics[width=2cm]{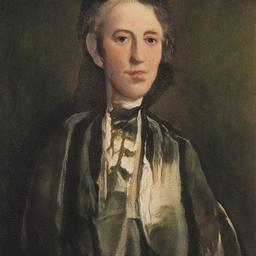}
		\includegraphics[width=2cm]{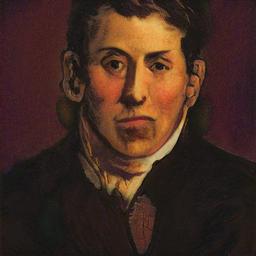}
		\includegraphics[width=2cm]{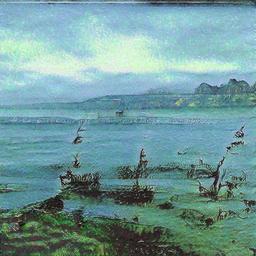}
		\includegraphics[width=2cm]{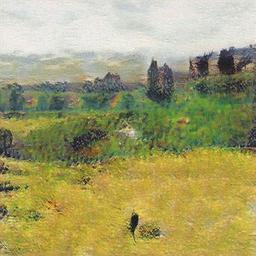}
		\includegraphics[width=2cm]{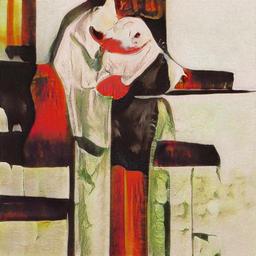}
	\end{center}
	\vspace{-0.5cm}
	\caption{Sample paintings generated by StyleGAN.}
	\label{fig:stylegan}
\end{figure}

As our baseline we use the TensorFlow implementation of StyleGAN \footnote{Available at \url{https://github.com/NVlabs/stylegan}} trained on the WikiArt painting dataset \cite{wikiart}.
As seen in \Cref{fig:stylegan}, StyleGAN produces very high-quality images, at times even indistinguishable from real paintings.
We do not expect to surpass StyleGAN with regards to artistic quality.
On the other hand, StyleGAN is not a T2I model, instead it just generates arbitrary images that might be convincing but have nothing to do with the provided text prompt.
Therefore, we are confident that our generated images will match better to the given captions.

\subsection{SSA-GAN}

\begin{figure}[t]
	\begin{center}
		\includegraphics[height=6cm]{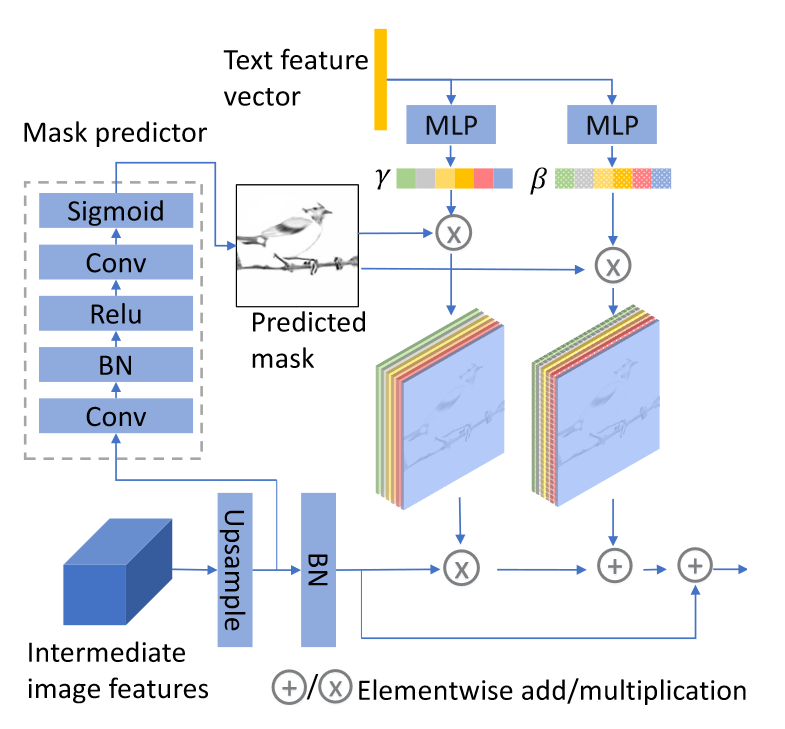}
	\end{center}
	\vspace{-0.5cm}
	\caption{Illustration of masked affine transformation taken from \cite{ssagan}.}
	\label{fig:ssagan}
\end{figure}

Both of the approaches we investigate build on a T2I model.
For this, we chose the Semantic-Spatial Aware GAN (SSA-GAN) architecture \cite{ssagan} that improves on DF-GAN \cite{tao2020DFGAN} by masking the affine transformation such that only certain image regions are fused with the text features (see Figure \ref{fig:ssagan}).
This mask is predicted from the image features and trained jointly with the rest of the network without a specific loss.
SSA-GAN also reuses the discriminator from DF-GAN and employs the DAMSM loss from \cite{xu2018AttnGAN}.
The pre-trained text encoder is a bidirectional LSTM \cite{schuster1997BidirectionalRN} adopted from \cite{xu2018AttnGAN}.
Compared to other works, its weights are not fixed but instead fine-tuned together with the generator.
For further details on the architecture we refer to \cite{ssagan}.

For our experiments we use the PyTorch implementation of SSA-GAN provided by the authors.\footnote{Available at \url{https://github.com/wtliao/text2image}}

\subsection{Style transfer approach}

For our first approach we train SSA-GAN on real-world images.
Given a text prompt we can then generate a realistic image that fits to this prompt.
In a separate step, we apply neural style transfer \cite{gatys2015style} to this image to make it look like a painting.
This way, we can benefit from the large amount of captioned real-world images available that can be used for training and use existing, well working techniques to shift the domain to artworks later on.
Concretely, we employ an SSA-GAN model that has been trained on the COCO dataset \cite{coco} for 120 epochs.

\paragraph{\textnormal{\bfseries Fast style transfer}}
Instead of the original style transfer proposed by Gatys et al. \cite{gatys2015style} we use so called fast style transfer \cite{Johnson2016Perceptual} that can stylize images in real-time.
Considering the time frame of the project, the original version which needs to solve an optimization problem for each images would be to slow, because we need a large amount of images for our later evaluation.
The idea behind fast style transfer is that a specific style image is baked into the network that can then stylize with a single forward pass.
We selected 30 paintings of varying styles from the WikiArt \cite{wikiart} dataset, taken from the epochs baroque, impressionism, and post-impressionism, and trained a stylization model for each of them.
\Cref{fig:wikiart} shows a selection of the images we used.

\begin{figure}[t]
	\centering
	\begin{minipage}{.4\textwidth}
		\centering
		\includegraphics[height=1.5cm]{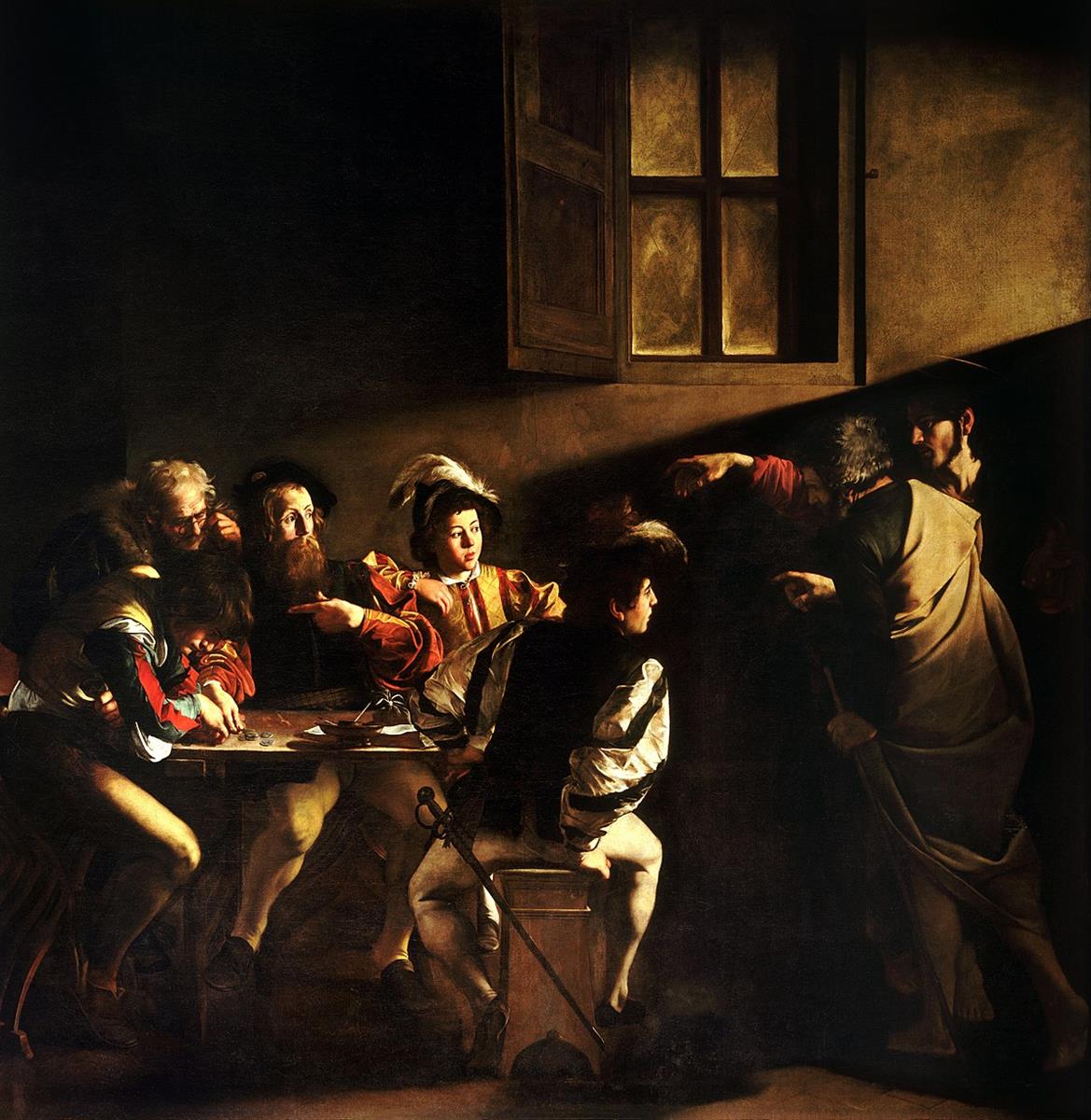}
		\includegraphics[height=1.5cm]{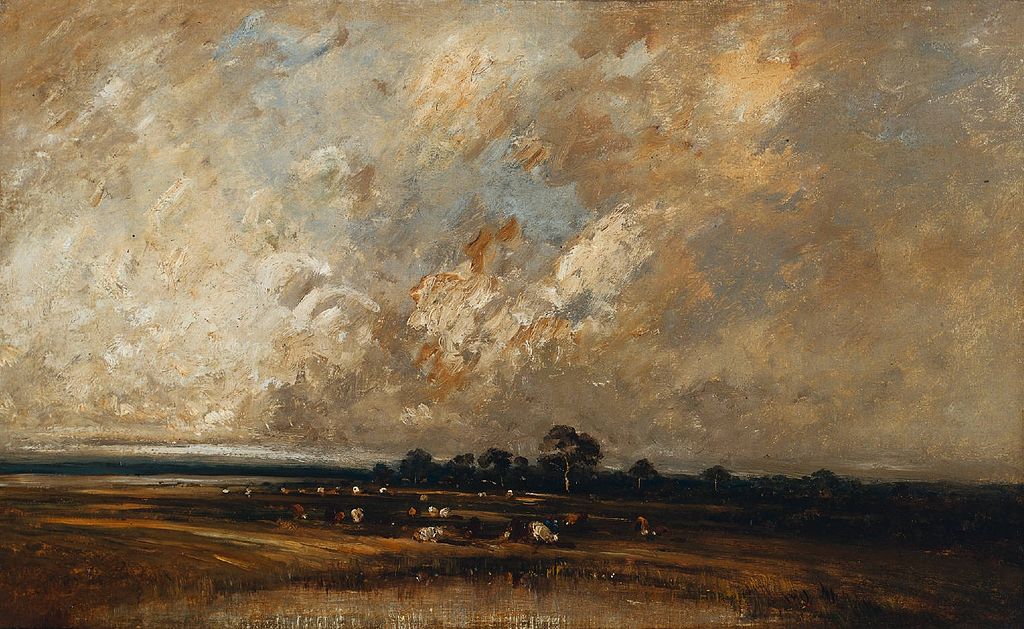}\\[2pt]
		\includegraphics[height=1.5cm]{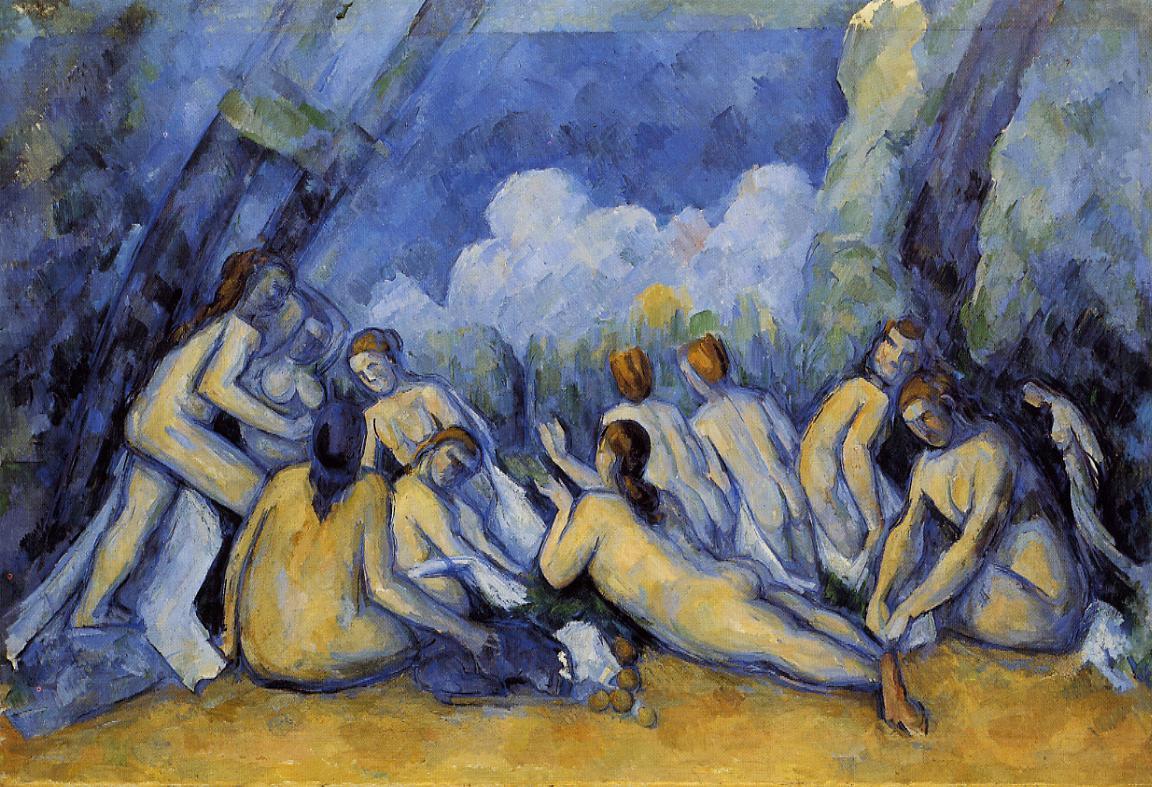}
		\includegraphics[height=1.5cm]{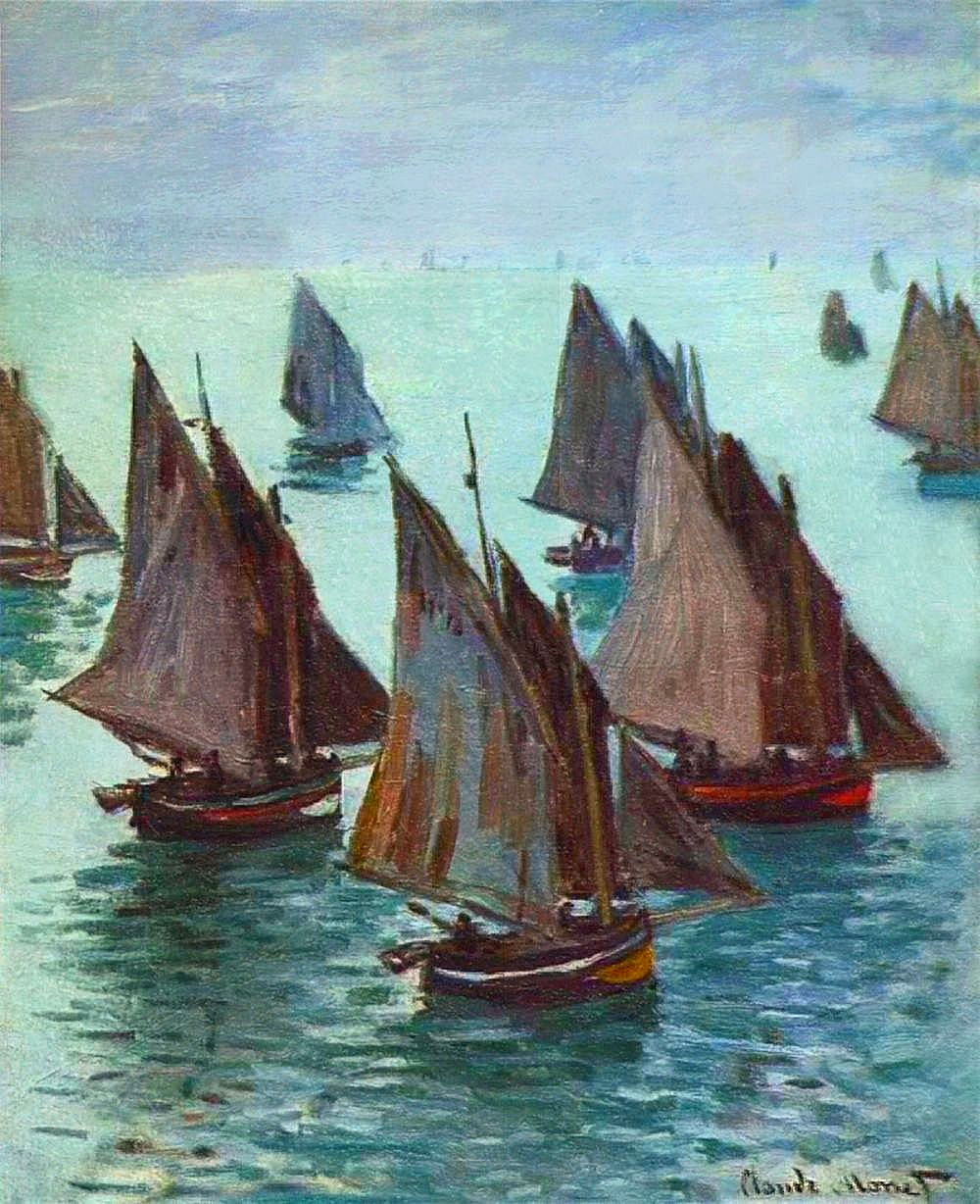}\\[2pt]
		\includegraphics[height=1.5cm]{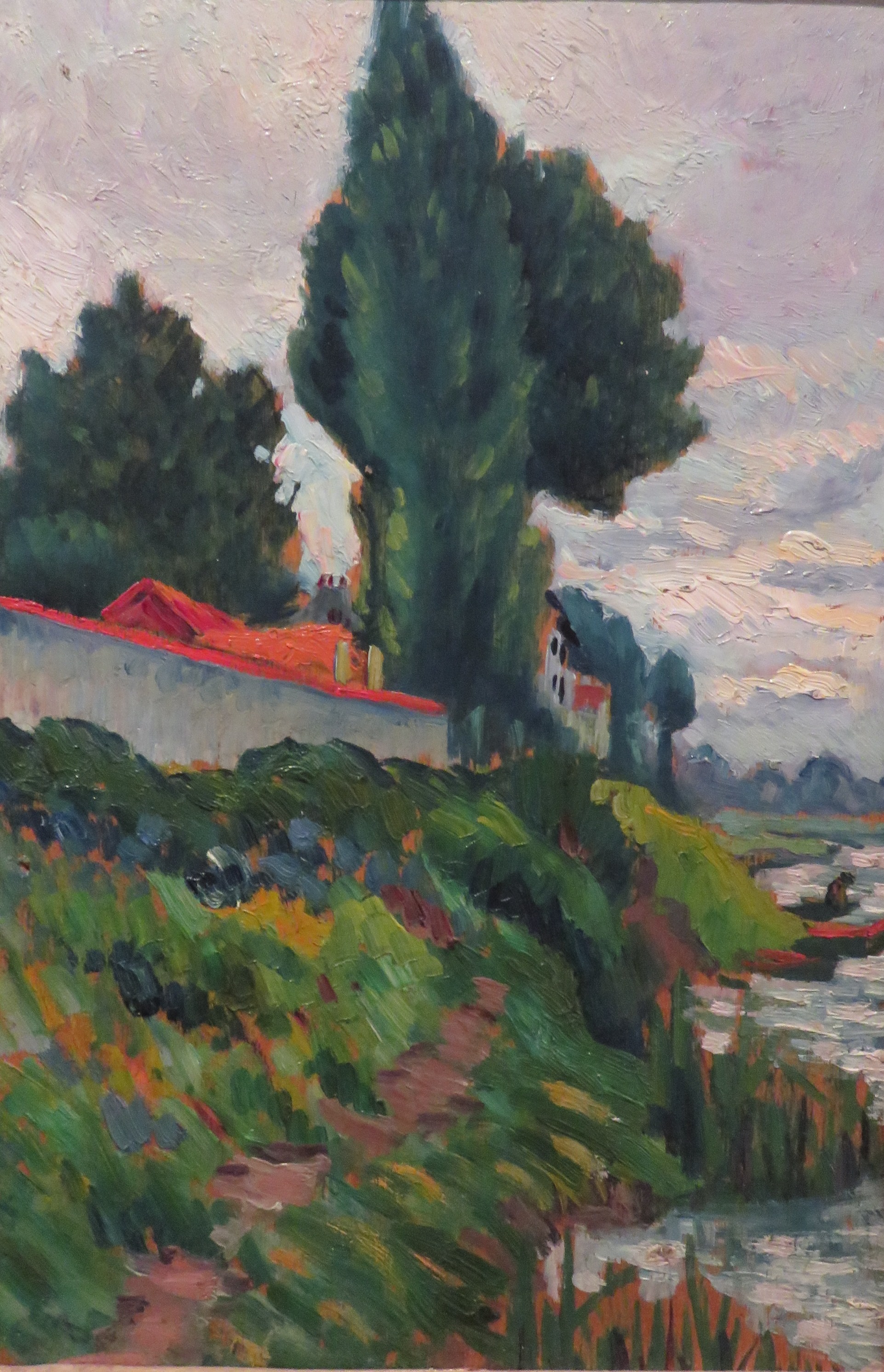}
		\includegraphics[height=1.5cm]{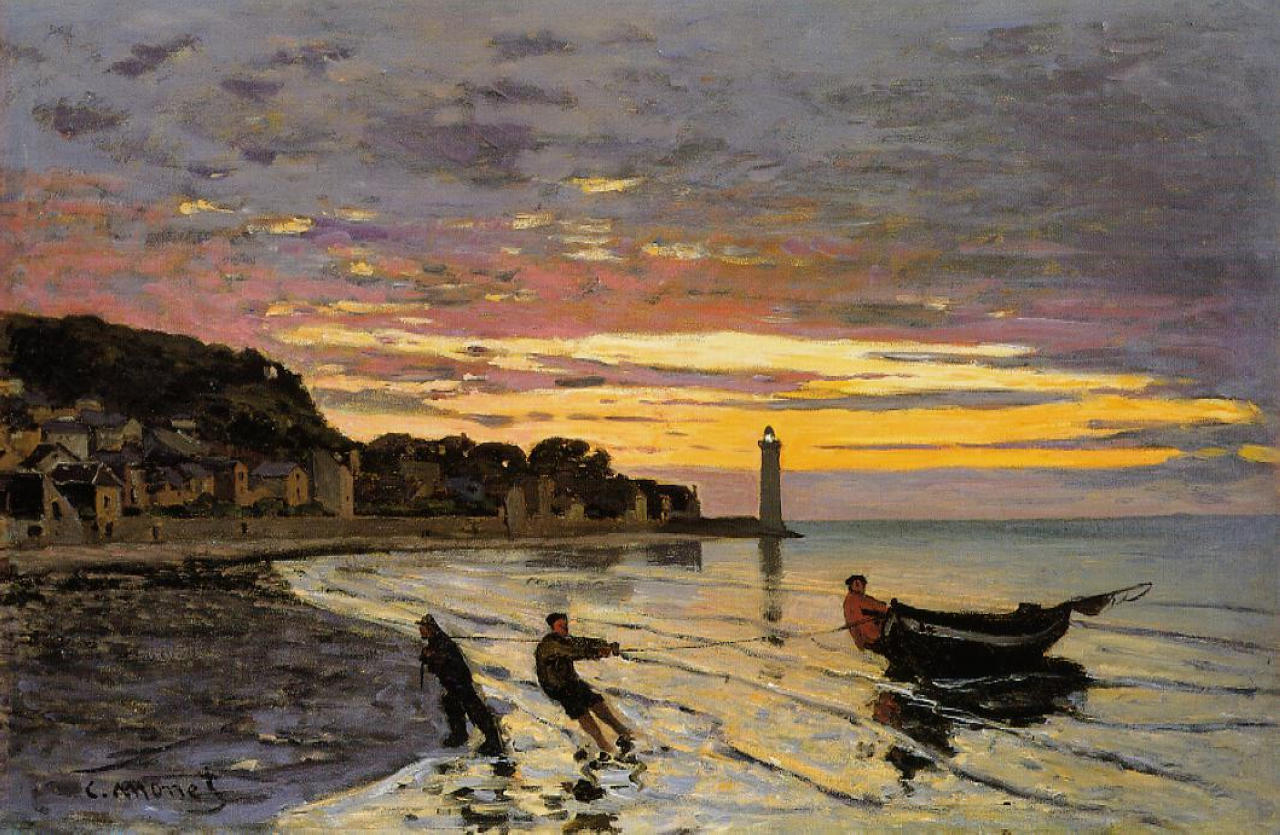}
		\caption{Selection of images from the WikiArt \cite{wikiart} dataset used for stylization.}
		\label{fig:wikiart}
	\end{minipage}
	~~
	\begin{minipage}{.5\textwidth}
		\centering
		\texttt{"A peaceful orange sunset over a mountain range"} \\[5pt]
		\scriptsize
		$\begin{array}{l} \includegraphics[height=1.5cm]{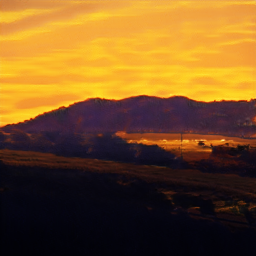} \end{array}
		+
		\begin{array}{l} \includegraphics[height=1.5cm]{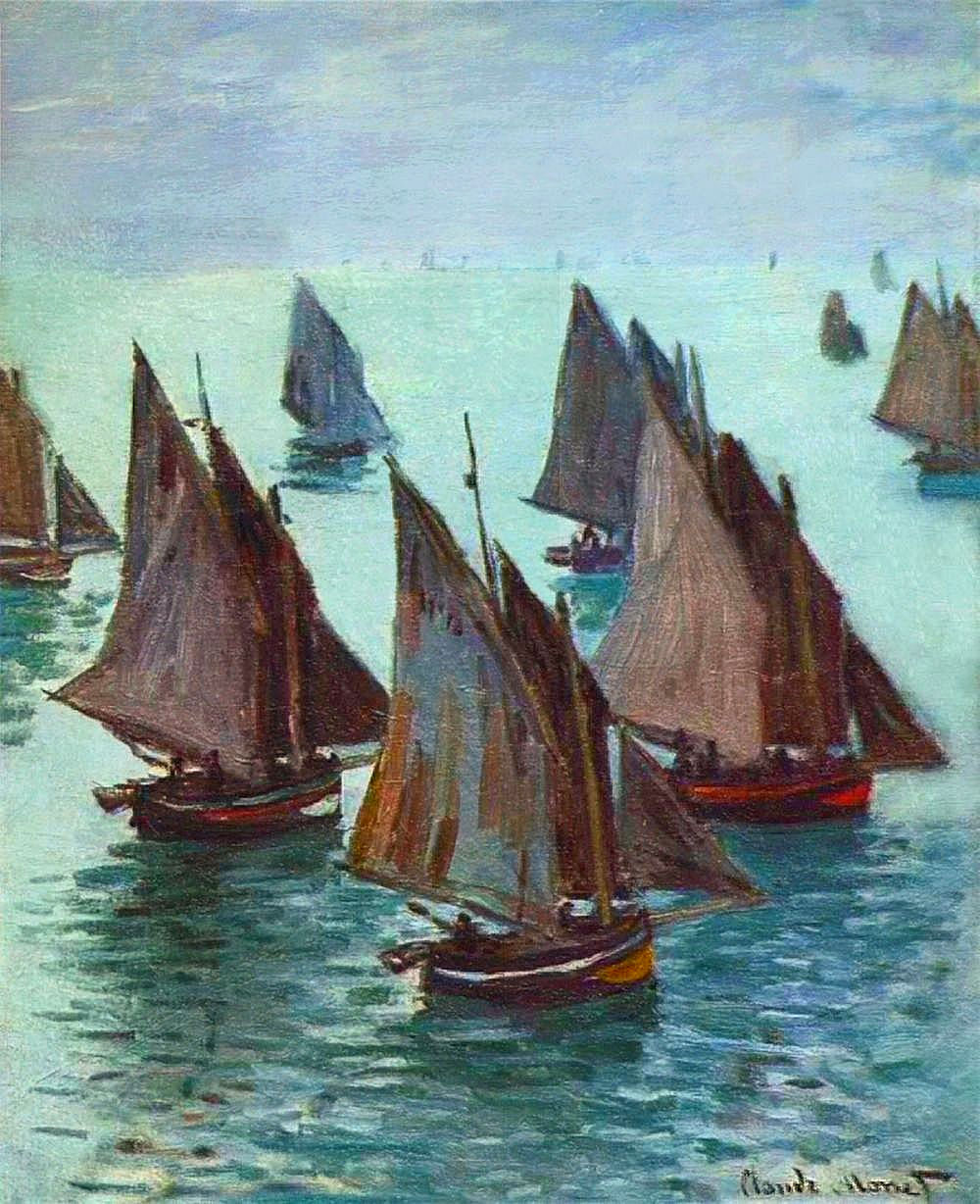} \end{array}
		=
		\begin{array}{l} \includegraphics[height=1.5cm]{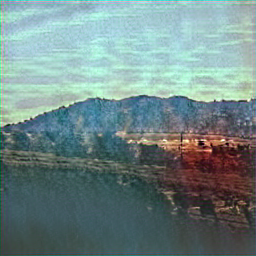} \end{array}$ \\
		$\begin{array}{l} \includegraphics[height=1.5cm]{img/styletransfer-hist1} \end{array}
		+
		\begin{array}{l} \includegraphics[height=1.5cm]{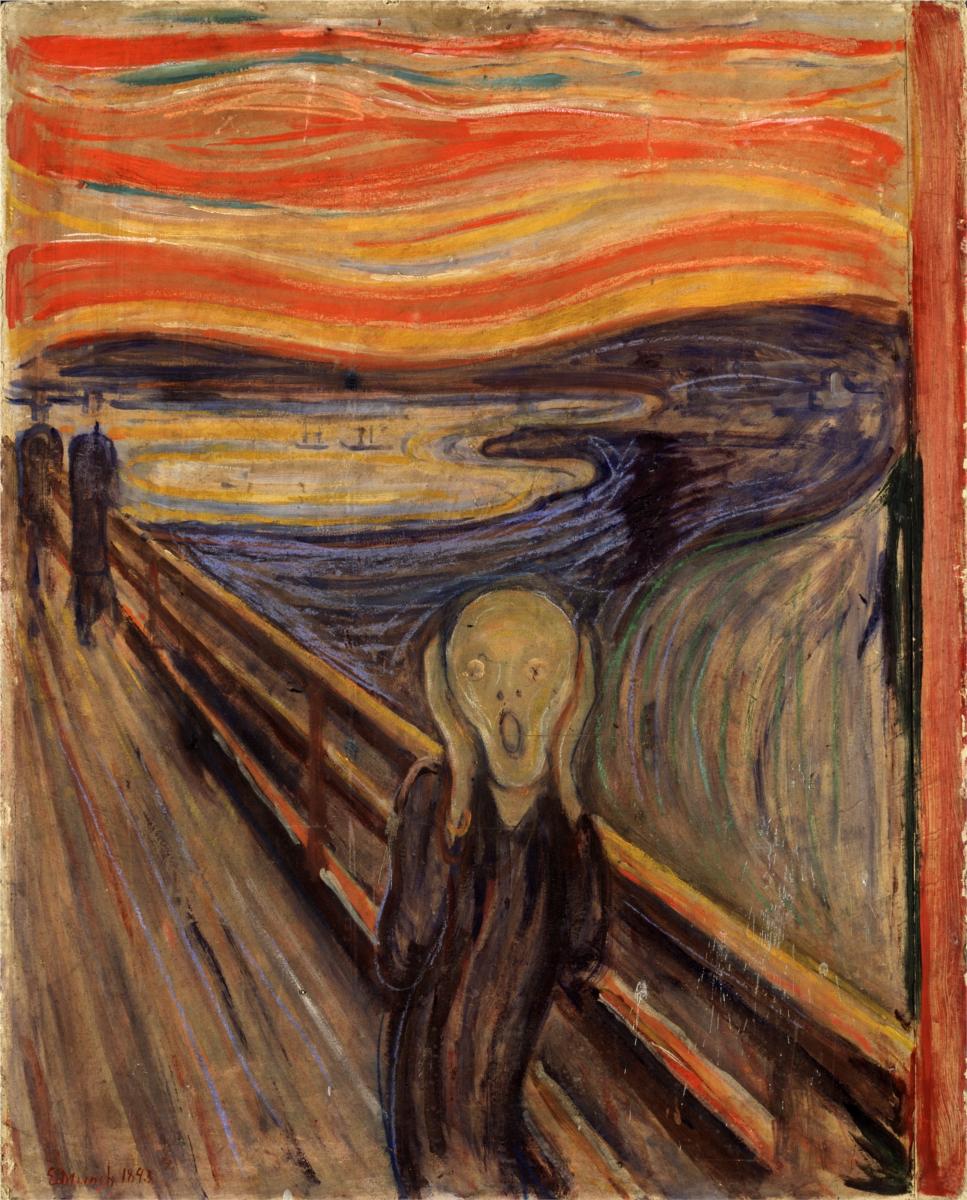} \end{array}
		=
		\begin{array}{l} \includegraphics[height=1.5cm]{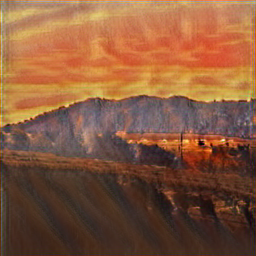} \end{array}$
		\captionof{figure}{SSA-GAN output combined with different style images. The \enquote{orange} gets lost if the wrong style is applied.}
		\label{fig:styletransfer-hist}
	\end{minipage}
\end{figure}

\paragraph{\textnormal{\bfseries Style image selection}}
Of course, after generating an image we still need to decide which of our paintings to choose to stylize with.
\Cref{fig:styletransfer-hist} illustrates that this is actually important, since the stylization can potentially alter the semantics of an image, in particular with regards to colours.
Therefore, we try to find the style painting from our collection whose colours match best with the generated image.
For this we employ the histogram intersection algorithm from \cite{swain1991histogram}.

\subsection{Fine-tuning approach}
For our second approach, we use the SSA-GAN again. Now that it is already trained on COCO, we can fine-tune it with the captioned painting data in order for the model to generate paintings directly instead of photorealistic images. The advantage of this is that the model already knows the 80 object classes that it learnt when training with the COCO dataset and by fine-tuning we hope that might be able to transfer this knowledge to the final paintings. When training from scratch it would be difficult to learn some concepts that do not typically occur in paintings (for example \enquote{keyboard}, \enquote{doughnut}, \enquote{microwave}) because there is no training data featuring those objects. Therefore, another benefit of fine-tuning is that we have overall more data to train the model with.

\paragraph{\textnormal{\bfseries{Captioning data}}}
Since we do not have access to any datasets which contain captioned paintings, we created our own by automatically generating captions for the WikiArt dataset using an image captioning model.
As image-to-text models have also been well researched and there are plenty of models available. 
We use the TensorFlow implementation of the NIC (Neural Image Caption) model \cite{showandtell} for this task\footnote{Available at \url{https://github.com/tensorflow/models/tree/archive/research/im2txt}\\
Checkpoints at \url{https://github.com/KranthiGV/Pretrained-Show-and-Tell-model}}. The NIC consists of a CNN that acts as an image encoder and an LSTM \cite{lstm} which generates the sentences. More details on that can be found in \cite{showandtell}.

Specifically, we captioned all the paintings of the same three epochs as before which amounts to around 23,000 pictures, thereof 4,000 images from the baroque, 13,000 from the impressionism and 6,000 from the post impressionism epoch. Some sample captions can be seen in \Cref{fig:caption}.

\begin{figure}[t]
	\begin{center}
		\begin{minipage}{.4\textwidth}
			\centering
			\includegraphics[height=3cm]{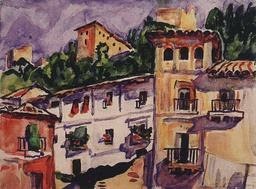}\\
			\texttt{A large clock tower towering over a city}
		\end{minipage}
		\begin{minipage}{.4\textwidth}
			\centering
			\includegraphics[height=3cm]{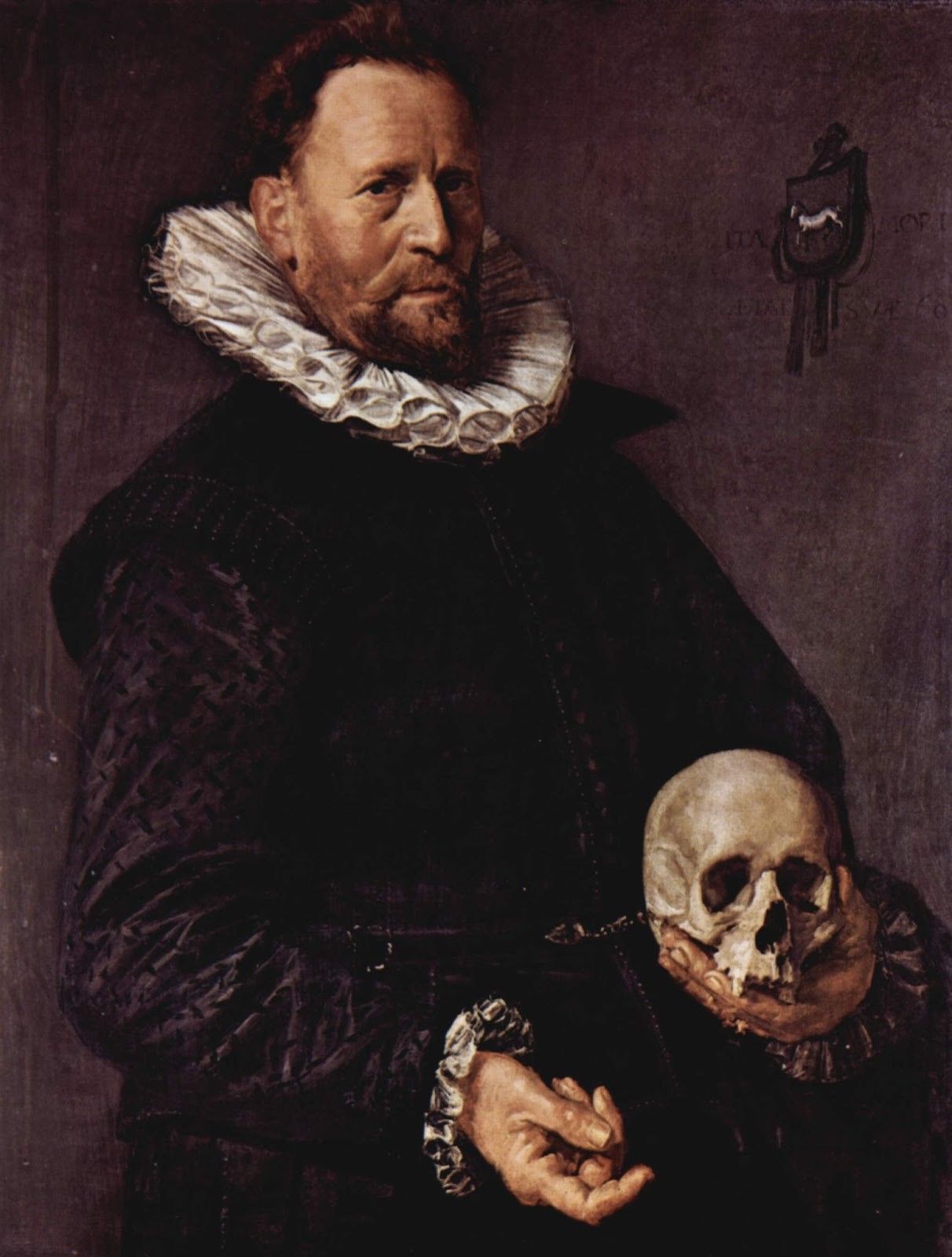}\\
			\texttt{A man in a suit and tie holding a teddy bear}
		\end{minipage}
	\end{center}
	\vspace{-0.5cm}
	\caption{Examples of captioned images with the NIC \cite{showandtell}}
	\label{fig:caption}
\end{figure}

We can observe, that the first image has been described well with identifying a city and a dominant building in it. However, in the image on the right the model misclassified the skull as a teddy bear. Overall, the results tend to go in the direction of the second example; this will be further discussed in \Cref{eval}.

\section{Experimental results}
\label{eval}

\begin{figure}
	\centering	
	\makebox[\textwidth][c]{\scriptsize 
		\renewcommand{\arraystretch}{1.25}
		\begin{tabular}{m{2cm}m{2cm}m{2cm}m{2cm}m{2cm}m{2cm}m{2cm}m{2cm}}
			&Initial image&Style transfer&Fine-tuning ~~~ (10 epochs)&Fine-tuning ~~~ (20 epochs)&Fine-tuning ~~~ (30 epochs)&Fine-tuning ~~~ (40 epochs)&Fine-tuning ~~~ (50 epochs)
			\\
			\texttt{A group of people walking down the road}
			&\includegraphics[width=2cm]{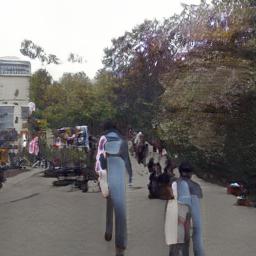}
			&\includegraphics[width=2cm]{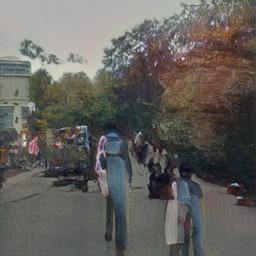}
			&\includegraphics[width=2cm]{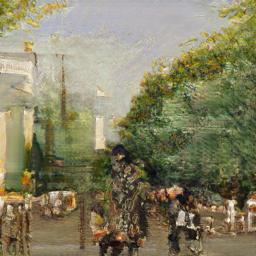}
			&\includegraphics[width=2cm]{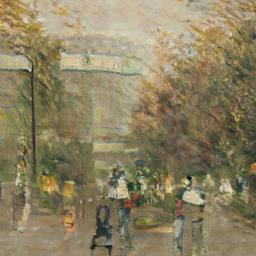}
			&\includegraphics[width=2cm]{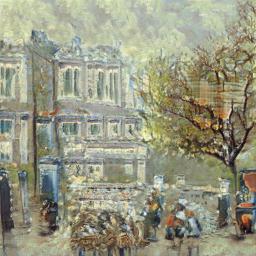}
			&\includegraphics[width=2cm]{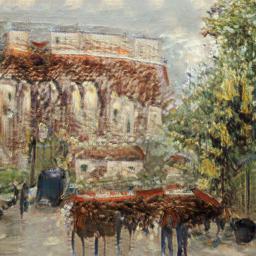}
			&\includegraphics[width=2cm]{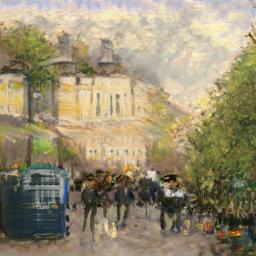}
			\\
			\texttt{There are cows walking on the grass}
			&\includegraphics[width=2cm]{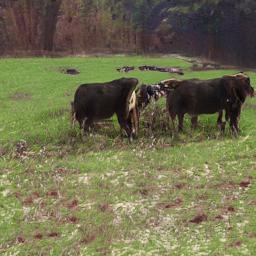}
			&\includegraphics[width=2cm]{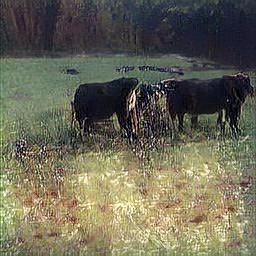}
			&\includegraphics[width=2cm]{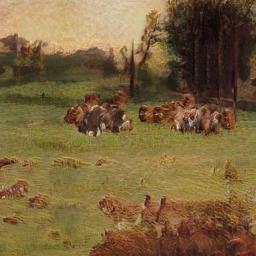}
			&\includegraphics[width=2cm]{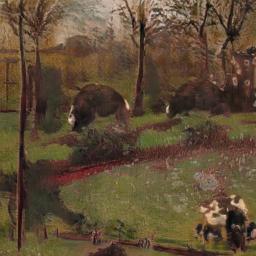}
			&\includegraphics[width=2cm]{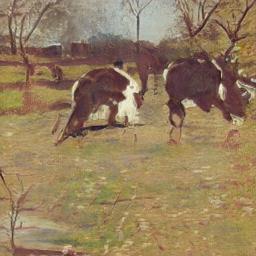}
			&\includegraphics[width=2cm]{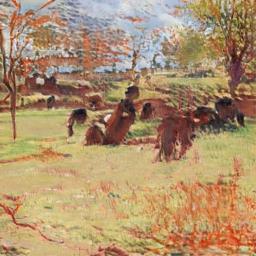}
			&\includegraphics[width=2cm]{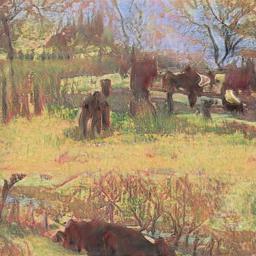}
			\\
			\texttt{Two men wearing neck ties around their heads}
			&\includegraphics[width=2cm]{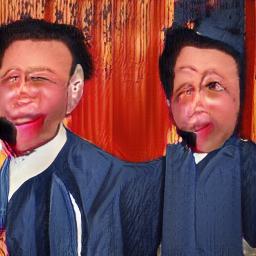}
			&\includegraphics[width=2cm]{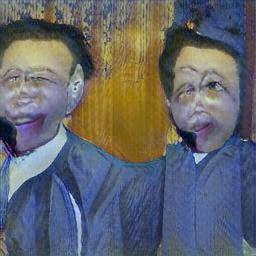}
			&\includegraphics[width=2cm]{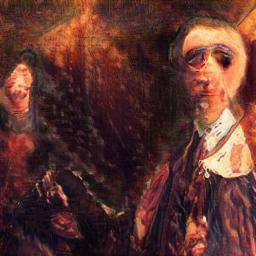}
			&\includegraphics[width=2cm]{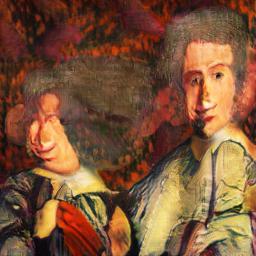}
			&\includegraphics[width=2cm]{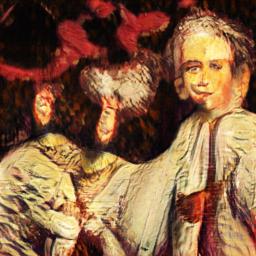}
			&\includegraphics[width=2cm]{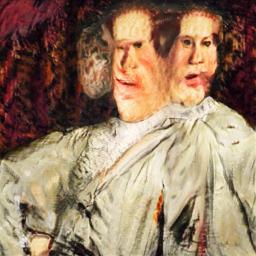}
			&\includegraphics[width=2cm]{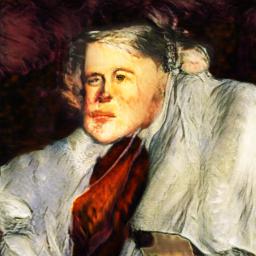}
			\\
			\texttt{A cozy living room with small potted plants and a flat screen}
			&\includegraphics[width=2cm]{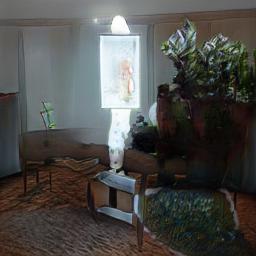}
			&\includegraphics[width=2cm]{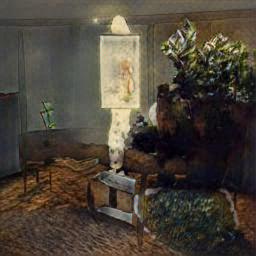}
			&\includegraphics[width=2cm]{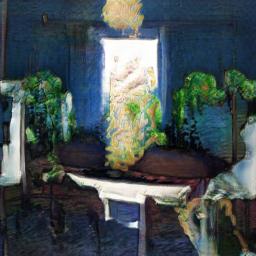}
			&\includegraphics[width=2cm]{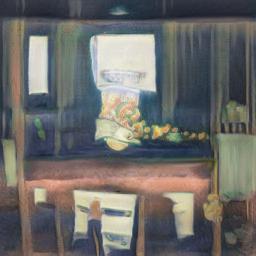}
			&\includegraphics[width=2cm]{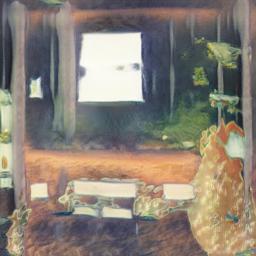}
			&\includegraphics[width=2cm]{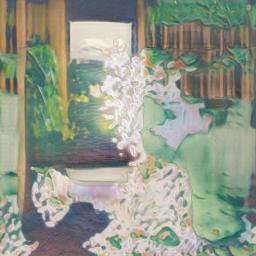}
			&\includegraphics[width=2cm]{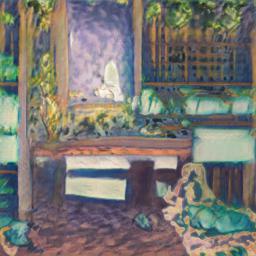}
			\\
			\texttt{A vase of red roses are sitting on a table}
			&\includegraphics[width=2cm]{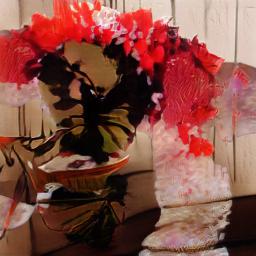}
			&\includegraphics[width=2cm]{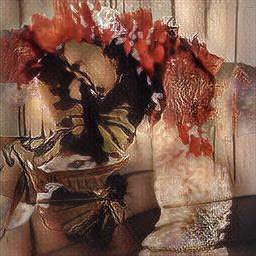}
			&\includegraphics[width=2cm]{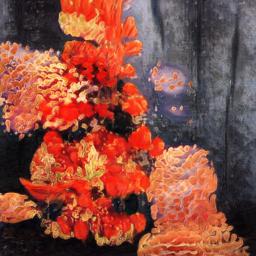}
			&\includegraphics[width=2cm]{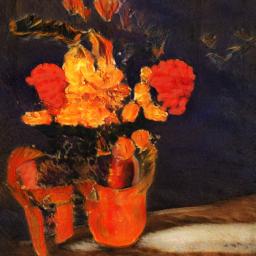}
			&\includegraphics[width=2cm]{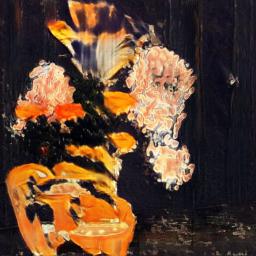}
			&\includegraphics[width=2cm]{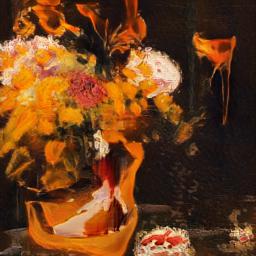}
			&\includegraphics[width=2cm]{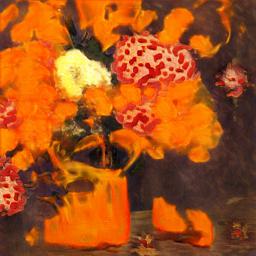}
			\\
			\texttt{A train on the tracks out in the country}
			&\includegraphics[width=2cm]{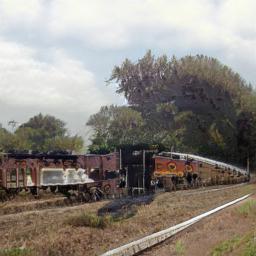}
			&\includegraphics[width=2cm]{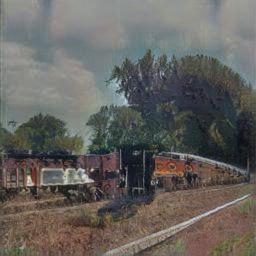}
			&\includegraphics[width=2cm]{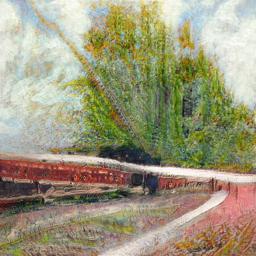}
			&\includegraphics[width=2cm]{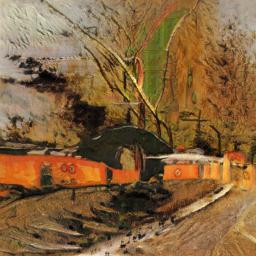}
			&\includegraphics[width=2cm]{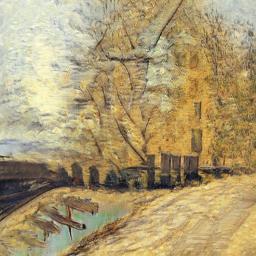}
			&\includegraphics[width=2cm]{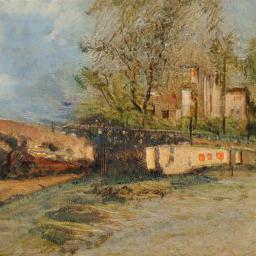}
			&\includegraphics[width=2cm]{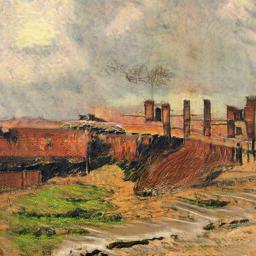}
			\\
			\texttt{A couple of giraffe standing on top of a dry grass field}
			&\includegraphics[width=2cm]{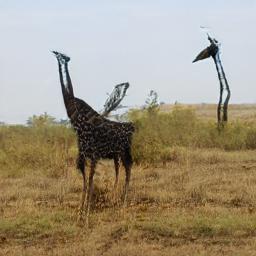}
			&\includegraphics[width=2cm]{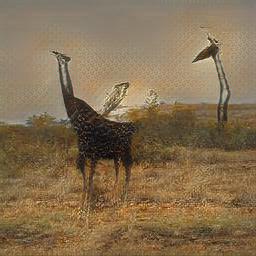}
			&\includegraphics[width=2cm]{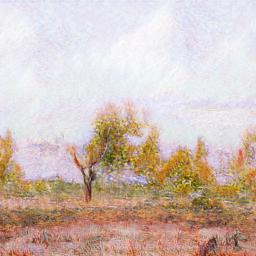}
			&\includegraphics[width=2cm]{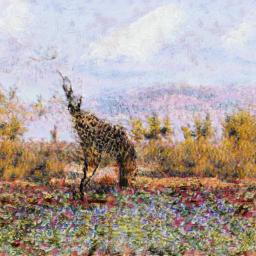}
			&\includegraphics[width=2cm]{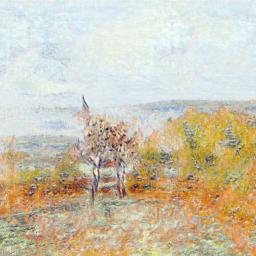}
			&\includegraphics[width=2cm]{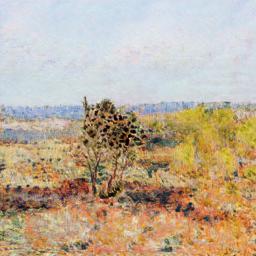}
			&\includegraphics[width=2cm]{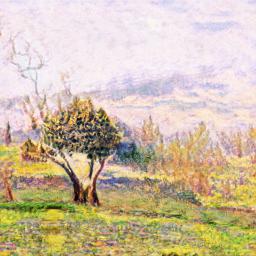}
			\\
			\texttt{There are many different donuts sitting on top of a table}
			&\includegraphics[width=2cm]{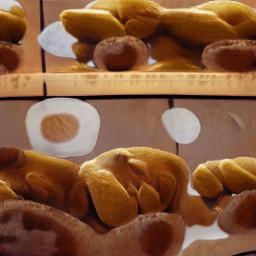}
			&\includegraphics[width=2cm]{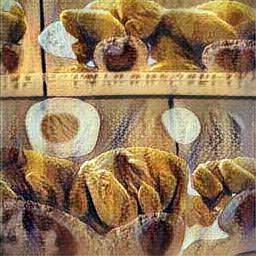}
			&\includegraphics[width=2cm]{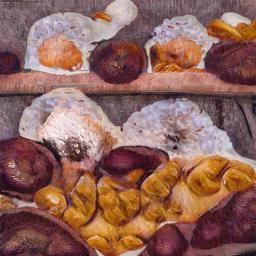}
			&\includegraphics[width=2cm]{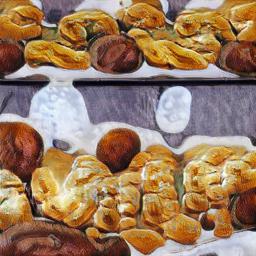}
			&\includegraphics[width=2cm]{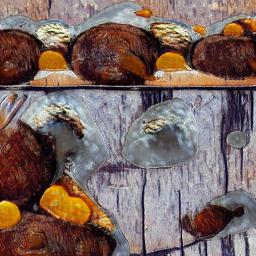}
			&\includegraphics[width=2cm]{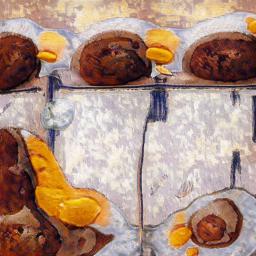}
			&\includegraphics[width=2cm]{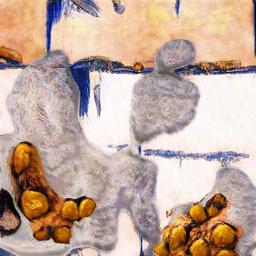}
			\\
			\texttt{A black stereo speaker near a computer monitor and mouse}
			&\includegraphics[width=2cm]{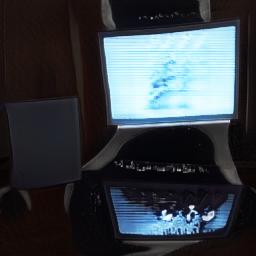}
			&\includegraphics[width=2cm]{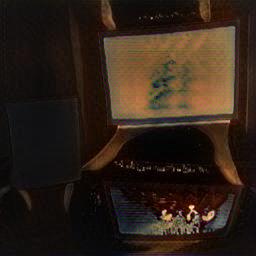}
			&\includegraphics[width=2cm]{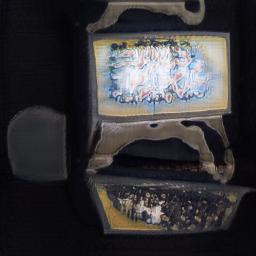}
			&\includegraphics[width=2cm]{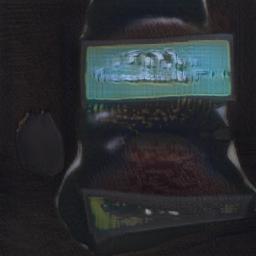}
			&\includegraphics[width=2cm]{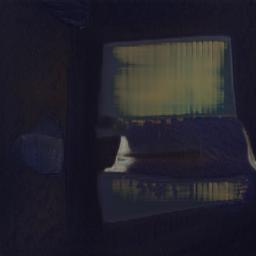}
			&\includegraphics[width=2cm]{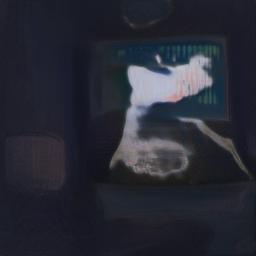}
			&\includegraphics[width=2cm]{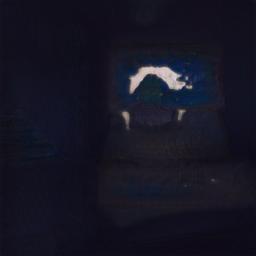}
		\end{tabular}
	}
	\caption{Qualitative comparison between the style transfer approach and different amounts of fine-tuning. The leftmost image shows the output of SSA-GAN purely trained on COCO.}
	\label{fig:results}
\end{figure}

We fine-tuned SSA-GAN for up to 50 epochs on the captioned WikiArt dataset.
Figure \ref{fig:results} shows examples of generated images using the style transfer and the fine-tuning approach for different numbers of epochs.
We observe that fine-tuning produces convincing nature scenes and even learns the style of baroque portraits used for training.
On the other hand, objects not present in the WikiArt paintings, like trains, giraffes, doughnuts or monitors, turn into landscapes and abstract shapes.
This indicates that the previous knowledge from training on COCO does not suffice to recover those objects after fine-tuning.
We can also see that the style-transfer approach stays closer to the content of the initial SSA-GAN output, making those  objects more recognizable.
However, we also notice that the stylization sometimes introduces artefacts or produces images that do not look like convincing paintings.

\subsection{Evaluation metrics}
Next, we quantitatively evaluate our two approaches.
It is comparably easy to evaluate the results of any supervised image classification or object detection model where the evaluation is quite straightforward. On the contrary, evaluating the generated results of a GAN is very challenging since it is not easy to quantify the realism of the generated image.

Some properties that we have targeted to evaluate our results with are as follows:
\begin{itemize}	
	\item \textbf{Fidelity and Diversity:} Measure how good the quality of our generated images is and how well they match the training data.
	\item \textbf{Captioning Accuracy:} Measure how relevant text captions are to the generated images.
	\item \textbf{Human Study:} Human Evaluation is important as it provides useful data on how the generated images are perceived by individuals. 	
\end{itemize}
In order to evaluate the results of our generated paintings on the basis of the above targeted attributes, we have decided on the followings metrics.

\begin{itemize}	
	\item \textbf{FID Scores (Fr\'echet Inception Distance)} \cite{Heusel2017GANsTB}: 
	Calculates the distance between feature vectors for real and generated images. This covers the fidelity and diversity properties.
	\item \textbf{SOA Scores (Semantic Object Accuracy)} \cite{soa}: 
	Focuses on individual objects and parts of an image and also takes the caption into consideration when evaluating an image. We have used these scores to evaluate caption accuracy.
	\item \textbf{Survey Application (Human Evaluation)}: 
	Many participants were asked to rate the images on the basis of persuasiveness, prettiness and captioning accuracy.	\\
\end{itemize}

\begin{table}[t]
	\centering	
	\begin{tabular}{|c|c|c|}
		\hline
		Model & FID $\downarrow$ & SOA $\uparrow$ \\
		\hline
		Baseline & 177.038 & - \\
		~ SSA-GAN trained on COCO ~ & - & 24.9 \% \\
		Style Transfer  & ~ 289.327 ~ & ~ 13.78 \% ~ \\
		Fine-tuned (10 epochs) & 241.252 & 8.28 \% \\
		Fine-tuned (20 epochs) & 234.593 & 7.02 \% \\
		Fine-tuned (30 epochs) & 252.014 & 6.29 \% \\
		Fine-tuned (40 epochs) & 259.254 & 5.54 \% \\
		Fine-tuned (50 epochs) & 258.391 & 4.85 \% \\
		\hline
	\end{tabular}
	\vspace{0.5cm}
	\caption{Comparison of FID and SOA with different models}
	\label{table:metrics}
\end{table}
\vspace{-1cm}

\subsection{FID scores}
The FID score (one of the most popular metrics used for evaluating GAN results) computes the Frech\'et Distance between the features distribution of the generated and real paintings. It is a measure of similarity between curves that takes into account the location and ordering of the points along the curves. 

\begin{quote}
	FID performs well in terms of discriminability, robustness and computational efficiency. [$\dots$] It has been shown that FID is consistent with human judgments and is more robust to noise. \cite{borji2018pros}
\end{quote} 
\Cref{table:metrics} shows the FID Scores we have achieved for the different approaches. A lower FID score indicates more realistic images that match the statistical properties of real paintings.\footnote{Note, that because of time constraints we were forced to use fewer samples than usual, so our scores are not directly comparable to other works.}

The baseline achieves the best score, as StyleGAN produces very good-looking and convincing paintings in contrast to our two approaches; especially style transfer suffers from the additional step of stylizing that introduced artefacts. The fine-tuning approach peakes at 20 epochs of training after which the values begin to stagnate at a worse level. This is the reason we chose those images for the human evaluation later on. 

\subsection{SOA scores}
Because FID does not actually take the image caption into account when evaluating the generated image, a generated painting could be judged with a good score even though it depicts something completely different than the image caption. SOA is a novel evaluation metric for T2I synthesis models that tries to tackle this issue by using an object detector to evaluate if a generated image contains the objects that it is supposed to contain according to the caption.

\Cref{table:metrics} shows the results of the SOA scores we have achieved. It indicates the percentage of images that actually contain the main object mentioned in the caption.\footnote{Note, that again we have fewer samples than in \cite{soa} so our results are not directly comparable to theirs.}

As a starting point, we can observe that our initial SSA-GAN model trained on COCO gets a score of roughly 25 percent, meaning this is an upper bound on what we can achieve using our two methods. The score for the style transfer approach is significantly worse which might be related to the fact that the object detector is not trained on paintings, therefore recognizing less objects. 

We can also see that the values for the fine-tuning approach decrease the longer we fine-tune the SSA-GAN which might be due to training with bad captions or the fact that the longer we fine-tune the more the SSA-GAN \enquote{forgets} about the classes trained before with the COCO dataset that are not contained in the paintings.

So, overall there is a trade-off between painting quality and caption accuracy with regard to the fine-tuning duration. For the human evaluation we therefore chose to use the model fine-tuned for 20 epochs since it provides a compromise between the two measures.

\subsection{Survey Application}
For our user study, we have developed a survey application where we have asked every user to rate 20 images on the basis of whether they thought the painting was real, and rating prettiness and caption accuracy on a scale of 1 (worst) to 6 (best). 
Of the 20 images, five were randomly taken from each style transfer, fine-tuning, baseline and real paintings as a control. In all of those categories 2400 images were available to, in turn, be picked randomly from. As the baseline model does not consider the captions we display random ones for the respective images; for the real paintings from the WikiArt dataset we display our generated captions which also allows us to evaluate their quality.


\noindent A total of 1080 images were rated, the results of which can be seen in \Cref{table:human}.

\begin{table}
	\centering
	\begin{tabular}{|c|c|c|c|}
		\hline
		Model &  Paintings Perceived as Real & Avg. Pretty Score & Avg. Caption Accuracy\\
		\hline
		Real Paintings  & {93.33 \%} & {4.94} & {2.74}\\
		Baseline & {76.32 \% } & {4.36} & {2.59}\\
		Finetuned & {64.48 \% } & {3.92} & {3.50} \\
		Style Transfer & {52.82 \% } & {3.83} & {3.94}\\
		\hline
	\end{tabular}
	\vspace{0.5cm}
	\caption{Results of the human evaluation survey}
	\label{table:human}
\end{table}
\vspace{-0.5cm}

\noindent It can easily be seen that the generated images from the StyleGAN get closest to real images with regard to image quality. On the other hand it performs worst for caption accuracy which is expected because the captions are chosen randomly.

The next best-looking images are produced by the fine-tuning approach. But it falls behind style transfer in caption accuracy which in turn  produces the most unauthentic paintings. This is possibly due to the artefacts that are introduced by the stylisation.

Overall, it can be observed that these results match to the scores that we obtained from FID and SOA.

Also note, that our generated captions were almost rated as bad as the random ones of the baseline indicating significant shortcomings of our captioned dataset.

\section{Conclusion and Future work}
In the many examples we generated we have seen a few images that looked very pretty and matched the given text prompt. The human evaluation also shows that a not insignificant amount of people considered the generated paintings as real ones. However, a large amount of the images produced have several shortcomings.

For example, we already have discussed, that the style transfer approach is superior with regard to caption accuracy because there is a realistic image generated first and the style transfer does not change the objects in the picture. However, the images generated that way suffer from patterns and artefacts which overall lowers their resemblance to real artworks. 
In future work, one might use the original style transfer model here instead of the fast one which should yield better results. This was not possible in our case for time reasons.

If we take a look at our fine-tuning approach, we already saw that the images are conceived prettier and more often mistaken for real paintings than in the style transfer approach. But fine-tuning also has some drawbacks with the main one being that the output often does not match at all with the given input caption. 
Especially modern objects like computers, kitchens or keyboards cannot be recognized on the respective images which indicates that too much information of this classes gets lost during fine-tuning.
Another reason for the overall bad connection between input caption and output image might be the very low caption accuracy we achieved with the NIC model for the WikiArt dataset . One has to expect that fine-tuning with descriptions that do not match the pictures lowers the quality of the output in general.
However, in future work there might be several ways to tackle this issues. For one, we could use a better image captioning model or manually clean the data. Secondly, there is actually one dataset available containing captioned paintings, namely the Bam! dataset \cite{bam}. However, we were not given access to it and thus could not use it. Using this dataset could also significantly improve the performance of the fine-tuning approach.

%
%
%
\bibliographystyle{splncs04}
\bibliography{biblio}


\end{document}